\documentclass[
]{ceurart}

\usepackage{tikz}
\usepackage[normalem]{ulem}

\begin{document}

\copyrightyear{2023}
\copyrightclause{Copyright for this paper by its authors.
  Use permitted under Creative Commons License Attribution 4.0
  International (CC BY 4.0).}

\conference{CEUR Workshop Proceedings (CEUR-WS.org)}

\title{Multi-Agent Coordination for a Partially Observable and Dynamic Robot Soccer Environment with Limited Communication }

\author[1]{Daniele Affinita}[%
email=affinita.1885790@studenti.uniroma1.it,
]
\author[1]{Flavio Volpi$^*$}[%
email=volpi.1884040@studenti.uniroma1.it,
]
\author[1]{Valerio Spagnoli$^*$}[%
email=spagnoli.1887715@studenti.uniroma1.it,
]
\author[1]{Vincenzo Suriani}[%
orcid=0000-0003-1199-8358,
email=suriani@diag.uniroma1.it,
]

\author[1]{Daniele Nardi}[%
orcid=0000-0001-6606-200X,
email=nardi@diag.uniroma1.it,
]

\author[2]{Domenico D. Bloisi}[%
orcid=0000-0003-0339-8651,
email=domenico.bloisi@unint.eu,
]

\address[1]{Department of Computer, Control, and Management Engineering Antonio Ruberti, Sapienza University of Rome
  }
\address[2]{Faculty of Political Science and Sociopsychological Dynamics, UNINT University, Rome
  }

\def\thefootnote{*}\footnotetext{These authors contributed equally to this work}\def\thefootnote{\arabic{footnote}}
  
\begin{abstract}  
  RoboCup represents an International testbed for advancing research in AI and robotics, focusing on a definite goal: developing a robot team that can win against the human world soccer champion team by the year 2050. To achieve this goal, autonomous humanoid robots' coordination is crucial. This paper explores novel solutions within the RoboCup Standard Platform League (SPL), where a reduction in WiFi communication is imperative, leading to the development of new coordination paradigms. The SPL has experienced a substantial decrease in network packet rate, compelling the need for advanced coordination architectures to maintain optimal team functionality in dynamic environments. Inspired by market-based task assignment, we introduce a novel distributed coordination system to orchestrate autonomous robots' actions efficiently in low communication scenarios. This approach has been tested with NAO robots during official RoboCup competitions and in the SimRobot simulator, demonstrating a notable reduction in task overlaps in limited communication settings. 
\end{abstract}


\maketitle

\begin{figure}[h]
    \centering
    \includegraphics[width=0.80\textwidth]{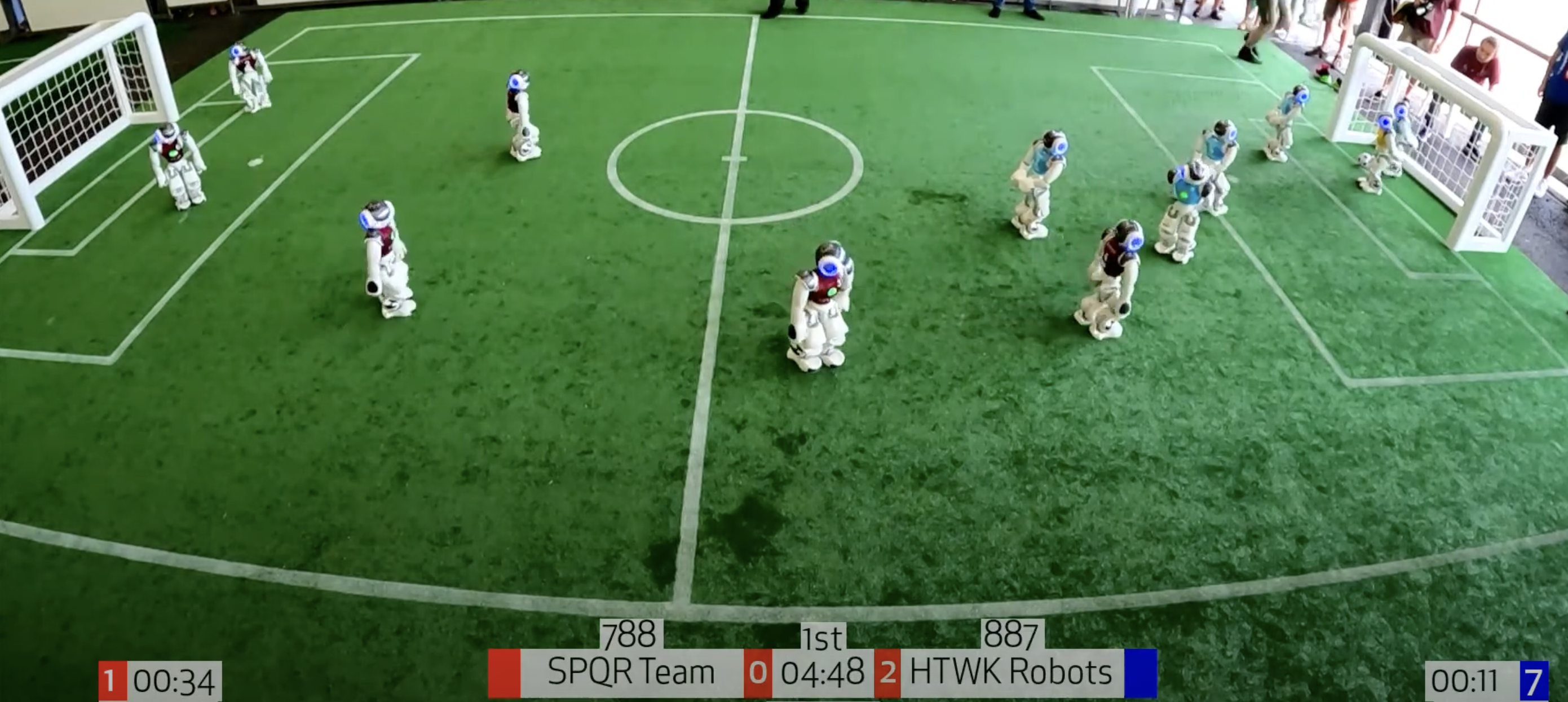}
    \caption{Frame from the quarter-finals of the RoboCup SPL 2023 between SPQR and HTWK teams. Above the names of the teams, the counters of the exchanged packets for each team are shown.}
    \label{fig:initial_screen}
\end{figure}

\section{Introduction}
Robocup is the world's largest robotics competition which aims to push the boundaries of research covering a wide range of topics.
Managing the coordination among a team of fully autonomous humanoid robots is a key aspect of dealing with the RoboCup 2050’s challenge, consisting of creating a team of fully autonomous humanoid robot soccer players able to win a soccer game complying with the rules of FIFA against the winner of the World Cup.

In the RoboCup Standard Platform League (SPL), the current trend is to rely less on WiFi communication, in order to push the boundaries of the robot's capabilities in managing the distributed task assignment problem in challenging conditions. Novel approaches, like gesture-based \cite{gesturecomm2019}, have been developed and tested, but wireless communication is still the main communication channel.

In particular, in the last few years, the network packet rate has been reduced from the original 5 packets per second per robot to a 1.200 total amount of packets per team per match. According to the rulebooks, from RoboCup 2019 to RoboCup 2022 the number of allowed packets per team has been reduced by 84\% \cite{rofer2022b}. In RoboCup 2023 the total number has been kept the same, but the number of playing robots per team increased from 5 to 7 (see Fig. \ref{fig:initial_screen}). This further reduced the amount of packets per robot. Meanwhile, the size of the single packet has been reduced to half of its size (now it is 128 bytes). This pushed teams to design new coordination paradigms and architectures.
To keep a team able to play a RoboCup match in a very dynamic and partially observable environment such as the RoboCup competition, it is needed to model the world representation, predict it when there are no updated data from the network and limited perceptions, and subsequently assign the tasks to the involved agents. 

The main contribution of this work is the definition of a new distributed coordination system, derived from the market-based task assignment for orchestrating the actions of multiple autonomous robots, ensuring their efficient performance even in setups with low communication rates. 
Our approach has been tested on real robots during competitions and in the SimRobot simulator\cite{rofer2021b} to evaluate the efficacy of our contributions, demonstrating how this approach can dramatically reduce the number of task overlaps in limited communication RoboCup matches.  

\section{Related Work} 

Coordinating a team of humanoid robots is a challenging task, especially in a very dynamic environment with hard constraints in the communication modalities. The RoboCup competition is one of the best testbeds for developing novel approaches, where a team of robots must cooperate effectively to compete in soccer matches. 
Within this competition, different leagues address the multi-agent coordination problem in distributed and centralized setups and using fully observable and partially observable environments, depending on the league\cite{robocupSurvey}. In SPL, the coordination can only be distributed and asynchronous.

An early approach for task allocation modeled as an asynchronous distributed system has been presented in \cite{farinelli2005task}. This system can either utilize each robot's perception or employ a token-passing mechanism to allocate tasks within the team. Lou et al. \cite{luo2011multi} propose an enhanced task allocation algorithm based on an auction system. They categorize potential tasks into subgroups and assign tasks to individual robots while ensuring precedence constraints are maintained. In Middle-size League (MSL), \cite{weigel2002cs} propose a task allocation strategy for the middle-size league of soccer based on utility estimations. They determine a set of preferred positions for the team based on the current situation and compute utility values to generate a reference pose set. 

In the 3D Simulation League, an advancement in robot coordination is introduced in\cite{macalpine2013positioning} involving a formation system algorithm. This algorithm computes a global world model shared among agents and locally evaluated. After each evaluation, robots broadcast their results. 

Additionally, other solutions to the challenge of coordinating heterogeneous robots are discussed in \cite{abeyruwan2012dynamic, vail2003multi}. These solutions rely on estimating the world state, mapping functions between robots and tasks, or mapping functions between robots and roles. 

To take advantage of the auction based-mechanism \cite{luo2011multi} with the estimation of the mapping functions between robots and roles\cite{weigel2002cs, vail2003multi}, and relying upon the local world model of the robot\cite{abeyruwan2012dynamic}, a unified approach, capable to manage also different playing contexts is presented in \cite{7759298}.  
In \cite{10.1007/978-3-031-28469-4_24}, to preserve the game capabilities of the robots, a dynamic sending approach is presented. 

To improve the placements of the robots on the field, some approaches rely on the Voronoi schema \cite{OffensivePlacementVoronoi2014, rofer2022b}. In contrast, our proposed method integrates these approaches by leveraging both distributed world knowledge and task-role assignments, but increasing the autonomy of the robots when no data are received from the teammates adding corrections on the robot positioning using the Voronoi diagram, as elaborated upon in the following section.

The proposed method creates a fully distributed market-based coordination system, inspired by the one proposed in \cite{7759298}, that leverages both distributed world knowledge and task-role assignment, and integrates a correction mechanism on the robot positioning using a Voronoi diagram, which allows improving the robots' autonomy in low-connection scenarios.

\section{Proposed Method} 
Our main contribution is the proposal of a market-based, distributed approach for multi-agent coordination, when there is a lack of information for an extended period. This specific topic has been overlooked in previous works and research which mainly focus on a standard situation in which it is always possible to share information among the agents. However, in a real-world application, it may happen that robot communication is not always possible or is delayed, especially when the communication medium is the network. Our methodology focuses on addressing this particular situation, leveraging the prediction models to compensate for the limited information exchange among agents. 

\begin{figure}[t]
    \centering
    \includegraphics{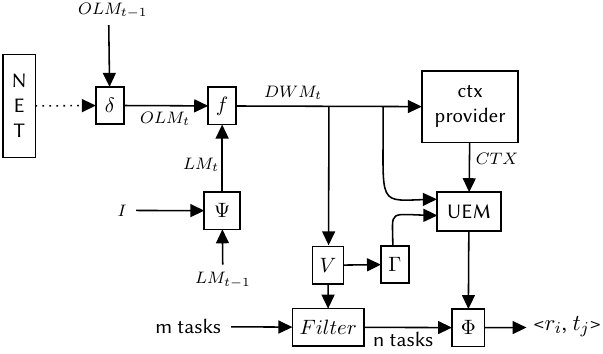}
    \caption{The overall architecture of \emph{DWM} and \emph{DTA}. The input is represented by a network event. If the event does not occur, prediction models probabilistically extend the previously estimated models. Then all local models are merged into \emph{DWM}, used to select the most valuable context and assign utility values to each <robot, task> pair. Finally, the optimal configuration \emph{V} is employed to match the number of roles to the number of available robots. Roles are then assigned to maximize cumulative utilities.}
    \label{fig:flowchart}
\end{figure}

In order to represent the operative scenario, we consider $M$ tasks, denoted as $T = \{\tau_1, \hdots, \tau_m\}$, and $N$ robots, denoted as $R = \{r_1, \hdots, r_n\}$, where in general $M > N$. Furthermore, we assume that we possess knowledge of an optimal robot placement configuration depending on the world state. In our study, a central theme that underscores the efficiency and effectiveness of our approach is the execution of both task assignment and world modeling in a distributed manner, without exchanging further information. In the context of the RoboCup domain, we consider robot roles as tasks; during the match, each robot must have a soccer role that defines its subtasks and goals.
The overall presented architecture is composed of several components, aiming to guarantee the execution in a challenging environment such as a RoboCup match where the teams can face unpredictable situations. The main components are represented by: \emph{a distributed world modeling, a position provider based on the Voronoi diagram, and a distributed task assignment procedure}, as depicted in Fig. \ref{fig:flowchart}. The distributed world modeling is achieved by fusing the information from all the robots, updated by a transition model that each robot adopts to keep a coherent representation of the world even under low packet rates circumstances. To easily propagate the information coming from the robot, the set of desirable positions is initially chosen using a Voronoi-based position generator. At the end of the procedure, each robot is capable of self-assigning a role and being aware of the teammates' roles without an explicit information exchange.

\subsection{Distributed World Model}

An essential prerequisite for an effective distributed task assignment algorithm is to have an accurate representation of the world. The local model of the world ($LM$) contains several components to represent the surrounding environment. The essential elements that contribute to world modeling include the obstacle model which incorporates the estimated poses of all robots and other rigid obstacles; a ball model, utilized for the estimation of the ball state (position and velocity); and a lines detector, employed to identify soccer lines within the field. Each component is derived from sensor data and is refined through the application of filtering techniques to mitigate perception errors. The inputs that affect the models are the robot's perceptions referred to as $I$ and the events $e$ sent from other robots through a common network.

We distinguish events that reflect the main situations in soccer matches. For example, an event is triggered when a robot detects a whistle from the referee. Another example of event triggering is when none of the team members have seen the ball for a while. In such a case, if a robot finds the ball, the context changes, and an event is triggered to notify the other agents.
It is important to notice that these events are in general not triggered at a specific rate, but they occur at irregular intervals, mirroring the dynamic nature of the real-world environment. Consequently, there exists the possibility of extended temporal gaps during which no event is sent through the network.

To address this limitation, we employ a function denoted as $\delta$ to update the $LM$ of other robots, referred to as $OLM_j$ for robot j. Specifically, this function merges the $OLM$ of the previous step with any received event, when available. In the absence of a received event, it uses a predictive model to compute a probabilistic model of the world. This allows us to avoid sending the whole $LM$ through the network obtaining a good estimation using only the available information.
\begin{equation} 
    OLM_{j,t} = \delta(OLM_{j,t-1}, e)
\end{equation}
At the same time, a function $\Psi$ updates the robot's local model $LM$ by incorporating input data received from the sensors into the previous local model.
\begin{equation} 
    LM_t = \Psi(LM_{t-1}, I)
\end{equation}
Both the local model update functions $\delta$ and $\psi$ involve predictive models which compensate for the absence of information, in case no network event is received. For example, a Gaussian Mixture Model (GMM) is employed to model the obstacles, a Kalman Filter for the ball preceptor, and odometry data for updating field elements and lines.
Having an updated version of the Local Model of every robot $\{OLM_{1,t}, \hdots, OLM_{n-1,t}, LM_t\}$, it is possible to reconstruct the \emph{Distributed World Model} ($DWM$) with a merging function $f$ which fuses the set of local models. 
\begin{equation} 
    DWM_t = f(OLM_{1,t}, \hdots, OLM_{n-1,t}, LM_t)
\end{equation}

\subsection{Distributed Task Assignment}
In our study, our primary objective is to enhance team coordination and strategic decision-making by adapting to the evolving configurations of the world. To achieve this, we introduce the contexts to represent various scenarios. Specifically, we rely on a module, the \emph{Context Provider}, which uses the information within the \emph{DWM} to dynamically select the most appropriate context (\emph{CTX}) from a predefined set. The context selection relies on a priority queue. Each context is linked to specific conditions. 
These contexts represent distinct situations in which a strategic adjustment becomes necessary.

At the same time, the information condensed in the \emph{DWM} is used as input to function $V$ to generate a set of desirable positions representing the optimal robot configuration at that moment. Notice that the configuration generated is role-independent, and each point within it does not represent an assignment to a specific robot but rather signifies a collection of potential waypoints. The points generated from $V$ have two purposes: filter $N$ out of $M$ tasks and further refine the Utility Estimation Matrix ($UEM$). 

The Utility Estimation Matrix represents the main data structure used to take into account the information from teammates and simulate task auctions locally. It is composed of N rows for robots and M columns for tasks, where the entry $(i,j)$ contains a non-negative number representing the utility. The utility is computed by considering several components that measure the effectiveness of a given \emph{DWM} with respect to a robot $i$ and a role $j$, so it quantifies how well robot $i$ can perform the task $j$. The final goal is to maximize the sum of all the assignments. The computation of $UEM$ is also influenced by the context selected by the context provider, allowing for adaptive role assignments based on the chosen strategy. The columns of the matrix are filtered using a module that compares the target of the roles with the waypoints derived from $V$. This filtering process transforms the matrix into an $N \times N$ square matrix, with an equal number of roles and agents. Finally, a function $\Phi$ provides the pairs $<r_i, t_j>$ from the filtered $UEM$:
\begin{equation}
    \Phi(UEM, tasks) \xrightarrow{} <r_i, t_j> \; \forall i,j
\end{equation}

The roles in the matrix are ordered by importance, meaning that a role in position $i$ has more priority than a role in position $j$ if $i < j$. The assignment process starts with the role in position 0 and assigns it to the robot associated with the row that maximizes its utility. Subsequently, we proceed to the next role in order of priority while considering the unassigned robots. This process allows each robot to simulate the potential assignments of other robots.
As the $DWM$ is probabilistically identical for all agents, each robot will reach an identical set of assignments.

\subsection{Voronoi Diagram} 

\begin{figure}
    \centering
    \includegraphics[width=1.0\columnwidth]{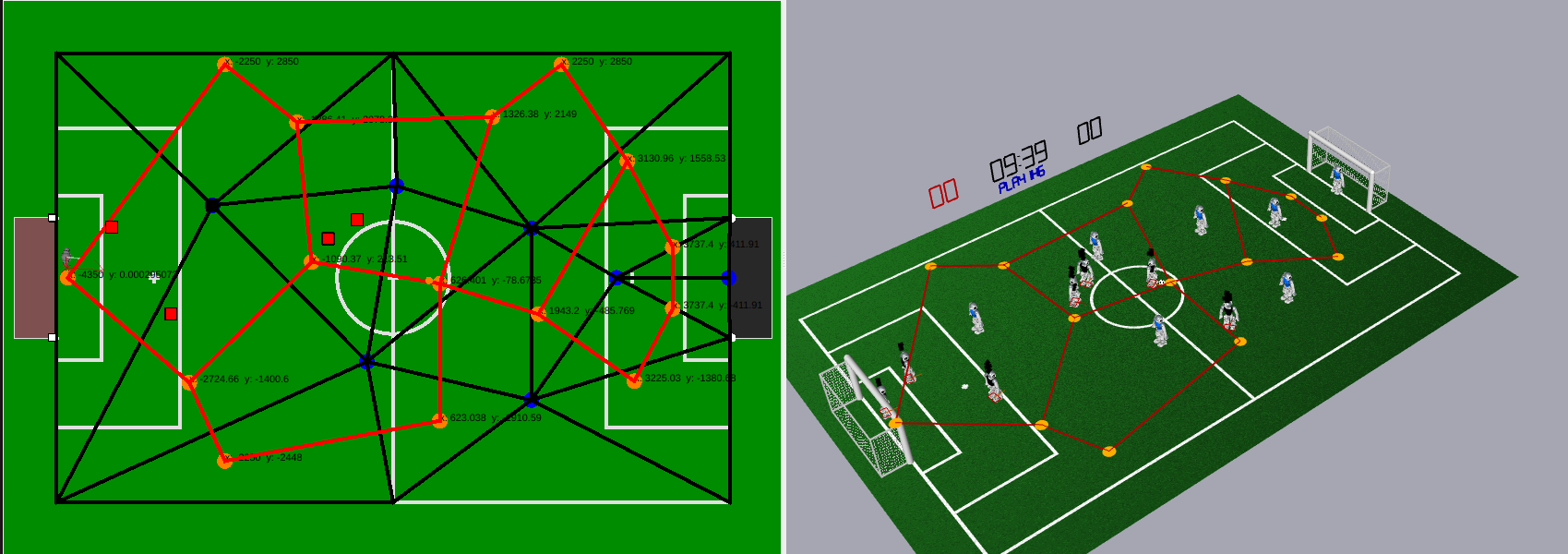}
    \caption{Voronoi graph in 2D and 3D field view. In the 2D view (left), blue points represent the opponent robots and black connections depict the Delaunay Triangulation, while red points are the Voronoi nodes and red links are the Voronoi edges.
    In the 3D view (right), just the Voronoi nodes and edges are shown.}
    \label{fig:voronoi}
\end{figure}

The function $V$ is a domain-specific optimal function that we assumed a priori, and which is used in the selection of the best N tasks and for the refinement of the UEM. The selection of the function is based on some precise aspects that are desired to maximize or minimize, according to the environment. 
In a RoboCup soccer field, where the coexistence of many robots in a limited space can create some issues in the evolution of the game, we are interested in maximizing the distances with adversarial robots. For this purpose, the function we chose is the Voronoi diagram (Fig.\ref{fig:voronoi}), which guarantees some advantages in the spatial disposition of the agents.

Given a set of $n$ points in the plane (called sites), the Voronoi diagram is the partition of the plane in polygons based on the distance to them. In particular, it ensures every point inside the same region is closer to its associated site than to the others. 
Formally, defined a metric distance $d$, we call $S=\{s_i| i=1,...,n\}$ the set of sites and $R=\{R_i|i=1,...,n \}$ the set of Voronoi regions, each one associated to the site $s_i$. Thus, taken a point $p$ of the plane:
\begin{equation}
    p \in R_i \iff d(p, s_i) \le d(p, s_j) \text{     } \forall j \ne i
\end{equation}
Every point $e$ such that $e \in R_i \land e \in R_j$ compose the Voronoi edge $E_{ij}$ between the polygons $R_i$ and $R_j$. So, the edge $E_{ij}$ is constituted by all the points that have the same distances with the sites $s_i$ and $s_j$, i.e.:
\begin{equation}
    E_{ij} = \{e | d(e, s_i) = d(e, s_j)\} \text{ with } e \in R_i \land e \in R_j
\end{equation}
Every point $v$ that belongs to at least three different Voronoi regions is called Voronoi node:
\begin{equation}
    v = R_i \cap R_j \cap R_k \cap ... \cap R_n
\end{equation}
In our case study, among all possible methods to build the graph, we decided to consider and construct it as the dual graph of the Delaunay triangulation, where the set of starting points is composed of all positions of opponent robots. The final Voronoi nodes and edges represent respectively the furthest points from the opponents and the optimal path to follow between two adjacent nodes. In other words, Voronoi nodes constitute the optimal positions for the team disposal.
The filtering process for the N out of M tasks is done through the proximity of the tasks to the nodes. In this way, we can ensure to pick the most suitable tasks for the environment evolution. 
The refinement of the UEM is performed by applying offsets to the tasks in their nearest node directions, displacing the task positions to the local optimal solution. 

\section{Experimental Results}
The system has been tested qualitatively during the last official RoboCup competition\footnote{https://2023.robocup.org/en/robocup-2023/}, and quantitatively in the SimRobot environment, simulating multiple matches.

To assess the performance of our approach, we employ the metric of \emph{multiple role periods}. Specifically, we have computed for each role, in each match, the total duration during which two or more robots assumed the same role simultaneously.
Since the \emph{striker} represents the most dynamic role with the highest priority, it best reflects coordination performance. 

We performed three sets of experiments, comparing the following approaches:
\begin{enumerate}
    \item multi-agent fixed-rate coordination that does not utilize events and Voronoi.
    \item multi-agent event-based coordination without Voronoi schema correction.
    \item multi-agent event-based coordination with Voronoi schema: the presented approach that includes the events and the novel Voronoi correction in the task assignment mechanism. 
\end{enumerate}

\begin{figure}[t]
    \centering
    \includegraphics[width=1.0\columnwidth]{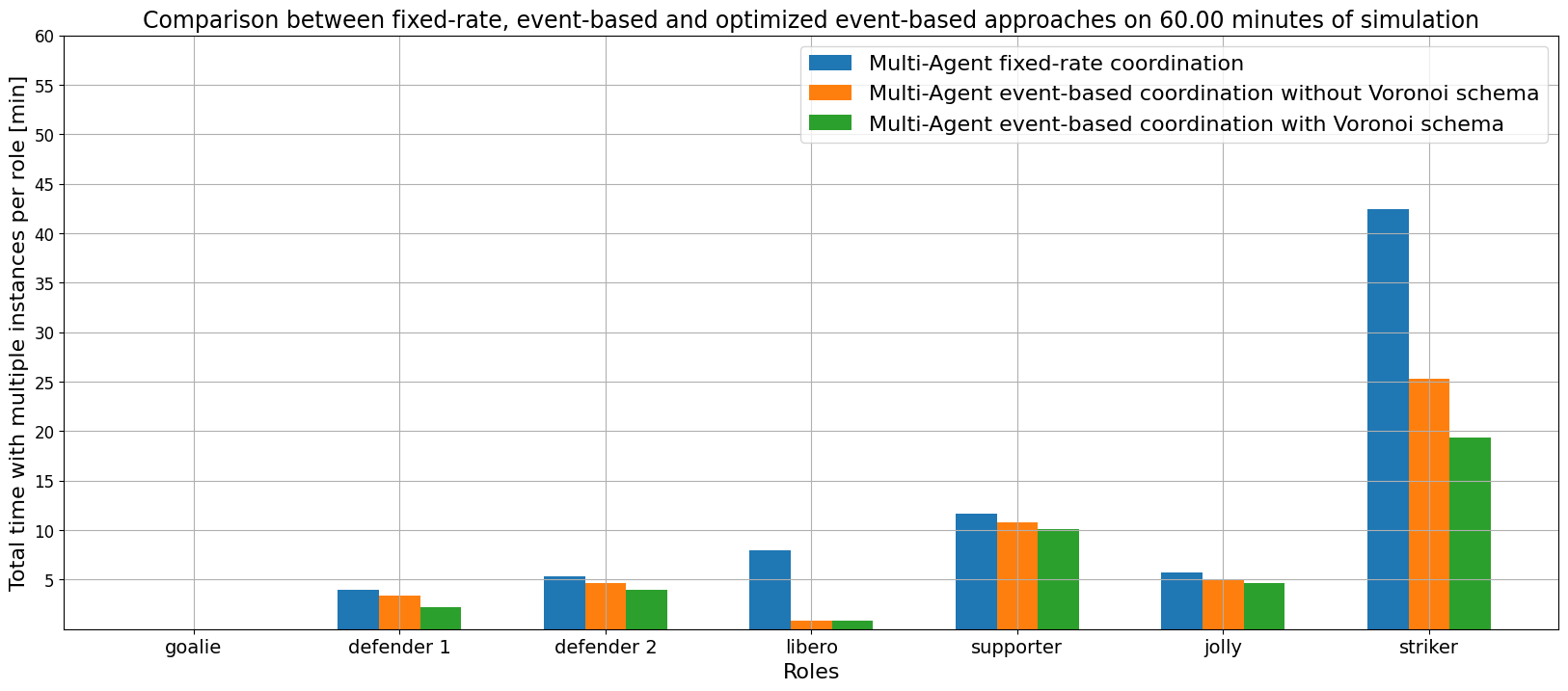}
    \caption{Role overlaps over time: for each role, the cumulative time (minutes) of role overlaps is shown. This demonstrates the improvements of the proposed approach (green) w.r.t. the baseline (blue).}
    \label{fig:results}
\end{figure}

The results, displayed in Figure \ref{fig:results}, show the cumulative role overlap duration. The x-axis represents the roles, while the y-axis represents the cumulative time. The total simulation duration is 60 minutes. The adoption of an event-based communication model allows for a more adaptive approach to environment changes compared to a fixed-interval rate, enabling the robots to communicate only when necessary. This is further improved using the Voronoi schema which obtained the best results in terms of overlapping time between roles (orange-green bars comparison). In fact, the Voronoi schema improves the coordination reducing role overlaps, by distributing the tasks of each role far from the tasks of the other roles, preserving effectiveness. 

\section{Conclusions}

In this study, we tacked the challenge of coordinating a team of fully autonomous humanoid robots participating in the RoboCup competition, in a low communication setup. The recent changes in SPL's rules, such as reduced network packet rates and an increased number of playing robots, prompted us to develop an innovative distributed coordination system based on market-based task assignments. 

Our system allows robots to model the world locally, propagate world predictions when network data is limited, and consequently efficiently assign tasks to team members. We adopted a market-based approach in which every robot simulates an auction locally assigning the available tasks to maximize the expected reward. We employed a Voronoi Graph to filter out additional roles to match the number of tasks with the number of available robots. Additionally, the Voronoi diagram has been also used for calculating a portion of the reward, contributing to the differentiation of the total reward. To address limited communication, we utilized prediction models to compensate for missing information from other agents, sending messages only when specific events occur.
Finally, we conducted extensive experiments, both in the real RoboCup environment and the SimRobot simulator, to assess our approach's performance.

The results clearly indicate that our approach effectively reduces task overlaps in low-communication scenarios, a critical factor in RoboCup matches. This research contributes significantly to the robotics field and RoboCup competition, offering a practical solution to the challenges posed by reduced communication rates in SPL. 



\bibliography{sample-ceur}



\end{document}